\documentclass[pdflatex,sn-mathphys-num]{sn-jnl}

\usepackage{graphicx}%
\usepackage{multirow}%
\usepackage{amsmath,amssymb,amsfonts}%
\usepackage{amsthm}%
\usepackage{mathrsfs}%
\usepackage[title]{appendix}%
\usepackage{xcolor}%
\usepackage{textcomp}%
\usepackage{manyfoot}%
\usepackage{booktabs}%
\usepackage{algorithm}%
\usepackage{algorithmicx}%
\usepackage{algpseudocode}%
\usepackage{listings}%
\usepackage{natbib}
\usepackage{float}

\theoremstyle{thmstyleone}%
\theoremstyle{thmstyletwo}%

\theoremstyle{thmstylethree}%

\begin{document}

\title[Article Title]{Hierarchical Multi-Label Contrastive Learning for Protein-Protein Interaction Prediction Across Organisms
}
\author[1,2]{\fnm{Shiyi} \sur{Liu}}\email{sliudd@connect.ust.hk}
\equalcont{First author.}

\author[1]{\fnm{Buwen} \sur{Liang}}\email{wliang1997@gmail.com}
\equalcont{Second author.}

\author[1]{\fnm{Yuetong} \sur{Fang}}\email{yfang870@connect.hkust-gz.edu.cn}
\equalcont{Third author.} 

\author[1]{\fnm{Zixuan} \sur{Jiang}}\email{zixuanjiang@hkust-gz.edu.cn}
\equalcont{Fourth author.}

\author*[1,2]{\fnm{Renjing} \sur{Xu}}\email{renjingxu@hkust-gz.edu.cn}
\equalcont{Corresponding author.}

\affil[1]{\orgdiv{Function Hub}, \orgname{The Hong Kong University of Science and Technology (Guangzhou)}, 
          \orgaddress{\city{Guangdong}, \postcode{511453}, \country{China}}}
\affil[2]{\orgname{The Hong Kong University of Science and Technology}, 
          \orgaddress{\street{Clear Water Bay}, \city{Kowloon}, \state{Hong Kong}, \country{China}}}

\abstract{
Recent advances in AI for science have highlighted the power of contrastive learning in bridging heterogeneous biological data modalities. Building on this paradigm, we propose \textbf{HIPPO} (\textbf{HI}erarchical \textbf{P}rotein-\textbf{P}rotein interaction prediction across \textbf{O}rganisms), a hierarchical contrastive framework for protein-protein interaction (PPI) prediction, where protein sequences and their hierarchical attributes are aligned through multi-tiered biological representation matching. The proposed approach incorporates hierarchical contrastive loss functions that emulate the structured relationship among functional classes of proteins. The framework adaptively incorporates domain and family knowledge through a data-driven penalty mechanism, enforcing consistency between the learned embedding space and the intrinsic hierarchy of protein functions. Experiments on benchmark datasets demonstrate that HIPPO achieves state-of-the-art performance, outperforming existing methods and showing robustness in low-data regimes. Notably, the model demonstrates strong zero-shot transferability to other species without retraining, enabling reliable PPI prediction and functional inference even in less-characterized or rare organisms where experimental data are limited. Further analysis reveals that hierarchical feature fusion is critical for capturing conserved interaction determinants, such as binding motifs and functional annotations. This work advances cross-species PPI prediction and provides a unified framework for interaction prediction in scenarios with sparse or imbalanced multi-species data.
}


\maketitle
\section{Introduction}\label{sec1}

Protein-protein interactions (PPIs) represent one of the most fundamental molecular processes in living systems and are pivotal targets for therapeutic development in disease treatment. Historically, PPIs have been investigated using experimental techniques such as yeast two-hybrid systems\cite{bruckner2009yeast}, co-immunoprecipitation assays \cite{kaboord2008isolation}, pull-down experiments \cite{aronheim1997isolation}, chemical cross-linking \cite{tan2016trifunctional}, and proximity-based labeling \cite{cho2020proximity}. However, the growing need to map vast interactomes, coupled with the prohibitively high costs and labor-intensive nature of experimental approaches, has rendered large-scale wet-lab characterization of PPIs increasingly impractical. To address this, computational strategies have gained prominence as scalable and cost-efficient alternatives, enabling systematic prediction and mapping of PPIs to accelerate the exploration of uncharted regions of the human interactome.

Recently, artificial intelligence (AI), and deep learning in particular, has emerged as a transformative force in computational biology\cite{singh2023contrastive,zhang2022application}. From sequence-based models to structure-aware neural architectures, AI-driven approaches have demonstrated remarkable performance in a wide range of tasks, including protein structure prediction\cite{jumper2021highly}, functional annotation\cite{kim2023functional}, and interaction inference\cite{lv2021learning}. A particularly promising class of methods in this domain is contrastive learning\cite{radford2021learning,oord2018representation,}, which belongs to the broader family of self-supervised learning techniques. By learning to discriminate between semantically similar and dissimilar pairs, contrastive models are capable of uncovering informative and generalizable representations, even in the absence of explicit labels. This paradigm has been successfully applied in computer vision\cite{ramesh2022hierarchical} and natural language processing\cite{radford2021learning}, and its adaptation to biological data has recently gained traction\cite{singh2023contrastive}.

In the context of PPI prediction, contrastive learning offers a compelling framework for capturing relationships between protein sequences and associated biological features. By training models to pull together representations of interacting protein pairs while pushing apart non-interacting ones, contrastive learning naturally aligns with the relational nature of the PPI prediction. Moreover, it enables the integration of diverse biological modalities-such as amino acid sequences, structural domains, and functional annotations-into a unified representation space. However, despite these advantages, existing contrastive learning approaches often treat protein representations in a flat, unstructured manner, neglecting the rich hierarchical organization of protein functions\cite{ashburner2000gene}. Proteins are classified into families, superfamilies, and domains, and are annotated with hierarchical ontologies such as Gene Ontology (GO), reflecting multiple levels of biological specificity and generalization. Ignoring these structured relationships can lead to suboptimal representations and limited biological interpretability.

Structure-based approaches are widely applied in general protein modeling due to their ability to capture spatial and functional features of well-folded domains. However, in the context of protein–protein interaction (PPI) prediction, their utility is limited by the prevalence of intrinsically disordered regions (IDRs), which do not form stable tertiary structures but instead exist as dynamic ensembles in solution\cite{Wright2015IDP}. As a result, sequence-based frameworks are often more suitable for modeling PPIs, as they can effectively account for interactions mediated by both ordered and disordered protein regions.

In this work, we propose a hierarchical contrastive learning framework tailored for protein-protein interaction prediction. Our method integrates protein sequence representations with structured biological knowledge through a multi-tiered contrastive objective. We introduce hierarchical contrastive losses that reflect the ontological structure of protein functions and employ a data-driven penalty mechanism to adaptively enforce alignment between learned embeddings and known functional hierarchies. This design enables the model to encode both fine-grained and abstract biological relationships, improving robustness, interpretability, and transferability. Extensive experiments on benchmark datasets demonstrate that our approach outperforms existing methods in both supervised and zero-shot settings, particularly in low-data regimes and cross-species generalization scenarios. Moreover, our ablation studies reveal the critical role of hierarchical feature integration in capturing conserved interaction determinants, such as binding motifs and shared functional domains. 

By unifying contrastive learning with hierarchical biological structures, this work advances the state-of-the-art in PPI prediction and contributes a general framework for leveraging biological hierarchies in machine learning. It opens new avenues for robust, scalable interaction modeling, particularly in under-characterized organisms and across diverse evolutionary contexts.

\section{Results}\label{sec2}

\subsection{HIPPO Introduces Joint Protein Sequences and Hierarchical Relationship Learning of Proteins and the PPI network}\label{subsec20}

Existing models \cite{lv2021learning, gao2023hierarchical, wu2024mape} primarily learn the protein interactome based on sequence or structural information. While these models are effective in capturing secondary and tertiary structural features, they often fall short in explicitly modeling diverse interaction properties and their relationships among proteins, such as functions, post-translational modifications (PTMs), and binding domains. To address this limitation, we propose a hierarchical multimodal pretraining framework designed to extract rich, structured protein representations for PPI prediction. This framework integrates heterogeneous information of protein from different sources and models both molecular content and biological context. Specifically, interacting proteins are first represented by their linear amino acid sequences, enriched with biophysical descriptors. Then, the interactions between proteins are captured using a protein interaction graph, in which proteins serve as nodes and their interactions as edges.
To comprehensively model protein-protein interactions, our framework is explicitly designed to integrate both hierarchical biological relationships and network-level interaction patterns. Sequence-derived features are obtained via a Protein Language Model (PLM), while an Annotation Language Model (ALM) encodes non-hierarchical annotations. Hierarchical information-including protein family and domain structures—is leveraged as supervisory labels in hierarchical contrastive learning, guiding the representation space to reflect biologically meaningful hierarchies. Simultaneously, a Graph Neural Network (GNN) is employed to model the global structure of the PPI network, ensuring that the learned embeddings capture not only local sequence and annotation information, but also topological dependencies among interacting proteins. Together, these components enable our model to preserve and exploit both hierarchical protein relationships and the structural complexity of PPI networks.
The architecture is explicitly designed to integrate hierarchical biological knowledge with the topological structure of protein–protein interaction (PPI) networks. It operates in two primary stages: (1) feature extraction and (2) hierarchical and network-aware representation learning.

In the feature extraction stage, two parallel encoding streams are employed:~a sequence encoder processes amino acid sequences, while an annotation encoder encodes non-hierarchical protein functions as binary vectors. Both encoders utilize six transformer-like blocks with skip connections and layer normalization to enhance expressiveness and stability. Local features are refined through a combination of 1D convolutions- both narrow and wide-which together capture context ranging from short to long-range dependencies. This setup yields rich representations that encompass both sequence-level and annotation-derived characteristics.

To incorporate biological hierarchy, we introduce a hierarchical multi-label contrastive learning objective. Hierarchical annotations-such as family and domain relationships-are used as supervisory signals, guiding the model to structure the embedding space according to known biological hierarchies. This encourages proteins with similar hierarchical attributes to cluster together, thereby embedding functional relationships directly into the latent space.

For PPI network modeling, the unified protein embeddings are used as initial node features in a PPI graph, where interactions are represented as edges. To capture complex topological patterns, we employ a Graph Isomorphism Network (GIN) consisting of three recursive blocks, each with a GIN layer, ReLU activation, and Batch Normalization. This enables the model to aggregate contextual information from neighboring proteins and encode network-dependent features critical for interaction prediction. In the final prediction stage, embeddings from two candidate proteins are concatenated and passed through a Multi-Layer Perceptron (MLP) classifier. This ensures that both hierarchical protein relationships and the PPI network context jointly contribute to robust and biologically meaningful PPI prediction.

\begin{figure}[htbp]
\centering
\includegraphics[width=1\textwidth]{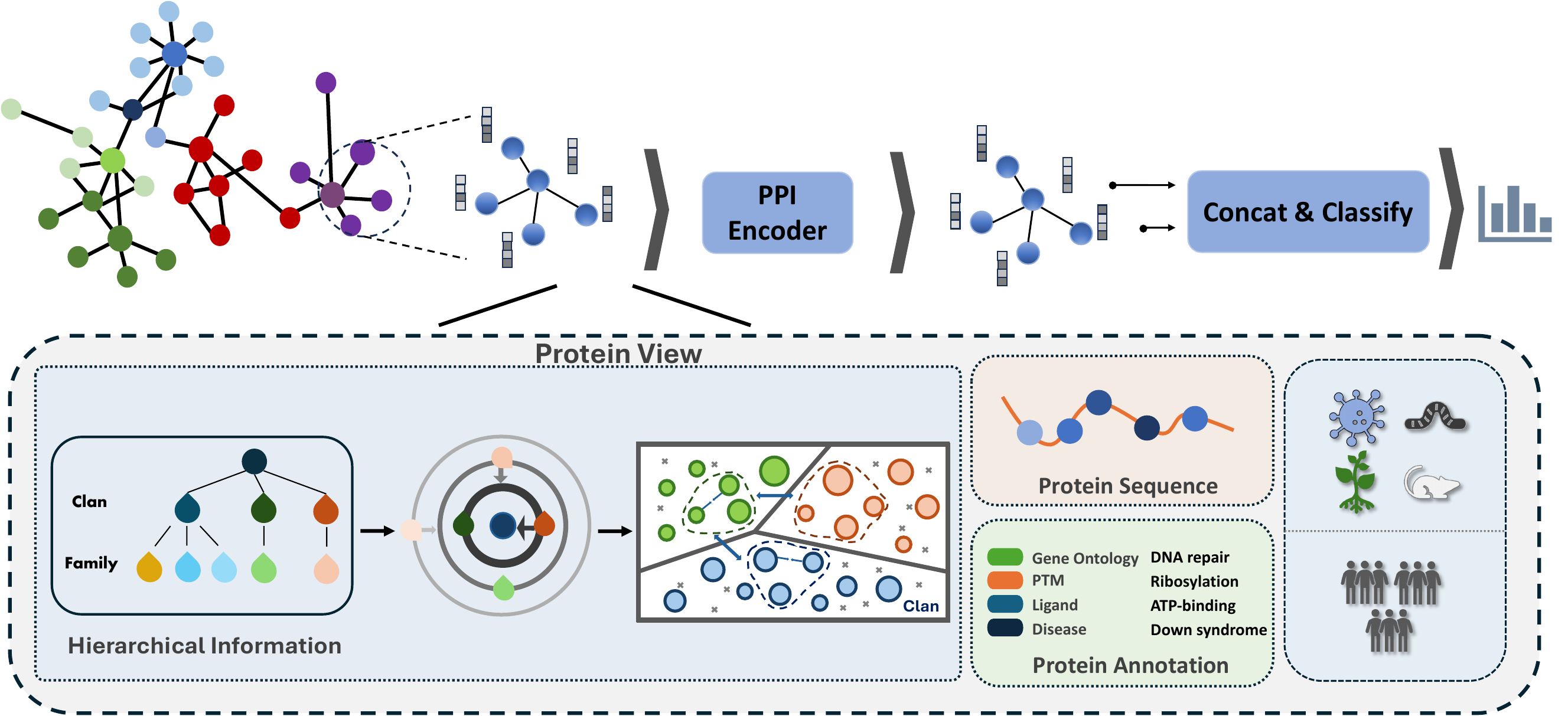}
\caption{Overview of the HIPPO architecture. The proposed framework include both the integration of protein sequence and hierarchical attributes to learn biologically meaningful representations and PPI network structure information. In the top panel~(PPI network), proteins are modeled as nodes, interactions are represented as edges, and the representations obtained from the bottom view are used as node features. Protein embeddings are refined through two trainable graph isomorphism blocks. For each query pair, the updated node representations are concatenated and passed through a linear classifier to learn and predict protein-protein correlations. In the bottom panel~(Protein View): Protein sequences are aligned with functional annotations including Gene Ontology terms, post-translational modifications(PTMs), ligand interactions, and disease associations. Hierarchical attributes-such as protein family hierarchy-is structured into a tree and embedded via contrastive learning that enforces locality in the representation space. Proteins with similar evolution trace are encouraged to cluster together, preserving hierarchical relationships in the latent space. }
\label{fig0}
\end{figure}

\subsection{HIPPO Achieves Robust Performance in Intra-species PPI prediction}\label{subsec21}

To validate the predictive power of the proposed model, we benchmarked its performance against leading methods from three perspectives: (1) overall predictive accuracy on three human datasets using depth-first search (DFS), breadth-first search (BFS), and random split strategies; (2) generalization across varying levels of prediction difficulty; and (3) classification accuracy across five distinct PPI types.

For a rigorous comparison, we froze the protein encoder to ensure consistent node feature extraction for the PPI graph, and performed PPI type prediction using a shared classifier layer for all models. Bar plots in Fig.~2A summarize the F1 scores of four models—our approach, GNN-PPI, ESM-2, and MASSA- on the SHS27k and SHS148k datasets, each evaluated under three different splitting strategies. The proposed method consistently achieves superior performance, improving the average micro-F1 by 2.9\%\ compared to the best baseline across all splits. Notably, under the DFS split, our model outperforms GNN-PPI by 10.9\%\ on SHS27k dataset,where partitioning is less favorable for learning from previously seen proteins. In contrast, the performance gap narrows considerably under the random split, with GNN-PPI slightly outperforming our model by 1.7\%\, indicating that both models are comparably effective when the data distribution is less stringent. All models demonstrate higher F1 scores on SHS148k compared to SHS27k, reflecting the greater redundancy and coverage of protein interactions in the larger dataset. Under the DFS split, our model surpassing GNN-PPI by 2.1\%\ . This advantage persists and becomes more pronounced under the BFS split, where our model has 3.5 \%\ improvement. Notably, model performance under the BFS split exhibits substantially higher variance across all methods. Under random split, the performance gap between methods is minimized, consistent with the less challenging nature of random partitioning. Therefore, our model delivers significantly improved predictive accuracy relative to leading baselines, underscoring its robustness in more realistic and challenging data scenarios.

To further assess generalization, we stratified protein–protein pairs into “easy” and “hard” categories: in the former, at least one interacting protein is seen during training, while in the latter, both proteins are unseen. As shown in Fig.~2B, all models achieve comparable F1 scores on easy pairs across both datasets and split strategies, indicating similar predictive capacity when at least partial training information is available. However, the differences between models become much more pronounced on hard pairs. While all methods experience a performance drop in this scenario, our model consistently outperforms the baselines and exhibits lower prediction variance. This advantage is particularly notable on the SHS27k dataset, where the gap between our method and others widens substantially for hard cases.

\begin{figure}[htbp]
\centering
\includegraphics[width=1\textwidth]{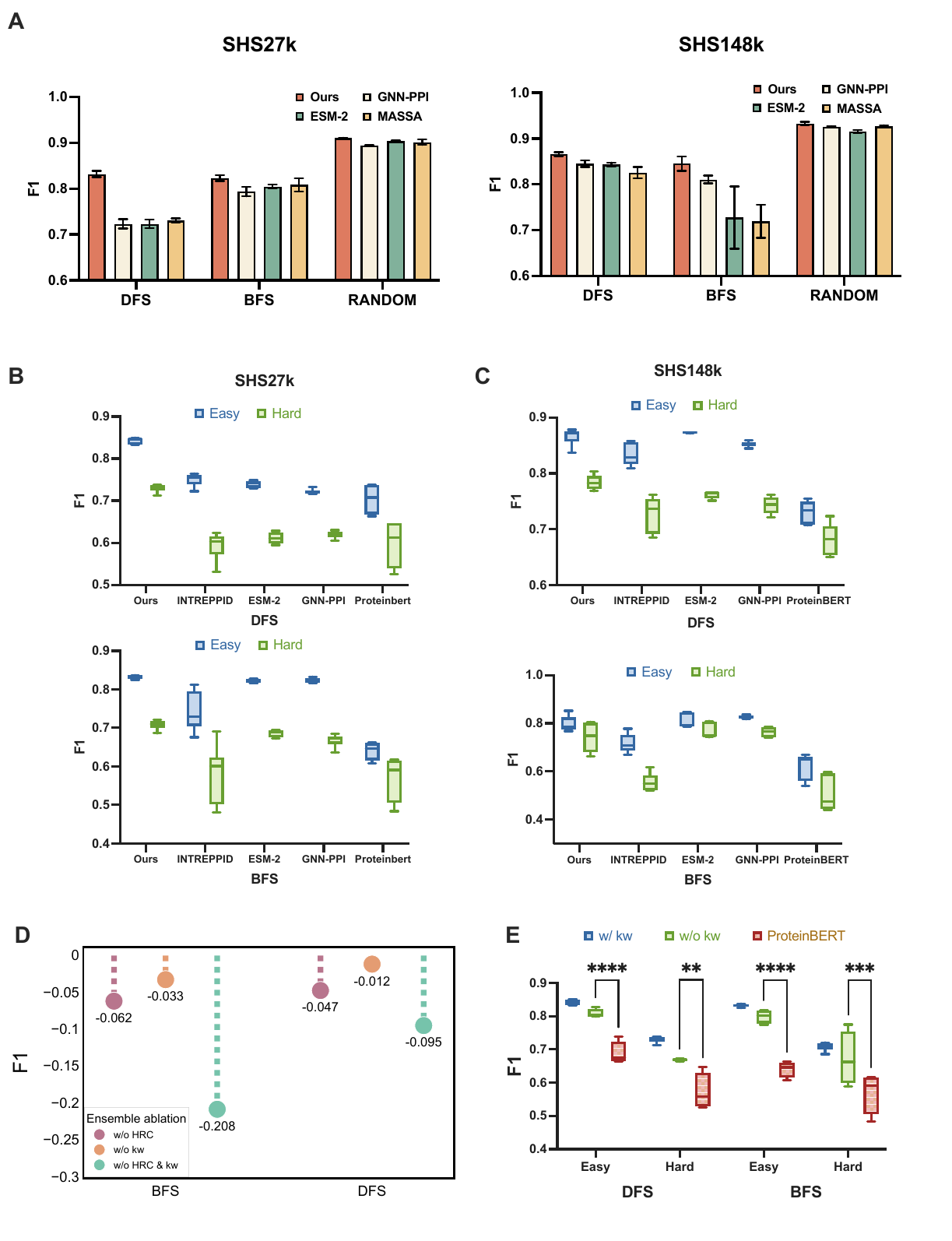}
\caption{Performance of HIPPO in predicting PPIs. (A)~Comparison of model performance across different datasets containing human proteins. Bar plots showing $\mathrm{F}_1$ scores scores of four models (Ours, GNN-PPI, ESM-2, and MASSA) evaluated on the SHS27k and SHS148k datasets using three atom selection strategies: depth-first search (DFS), breadth-first search (BFS), and random sampling. The proposed method consistently achieves superior or competitive performance across all settings. (B)~Comparative analysis with varying difficulty levels in SHS27k dataset. PPIs are categorized into easy(at least one interacting protein in the training sets) and hard(0). Box plots comparing $\mathrm{F}_1$ scores of various models on different difficulty of interaction subsets under both DFS and BFS sampling settings. (C)~Comparative analysis with varying difficulty levels in SHS148k dataset. (D)~Change of performance on SHS27k and SHS148k when excluding hierarchical attributes(HRC) and keyword annotations(kw). Scatter plots displaying the relative drop in performance ($\Delta \mathrm{F}_1$) resulting from ablating key components.  (D)~Comparison of model performance with keyword annotations (``w/ kw''), without keyword annotations (``w/o kw''), and the baseline ProteinBERT method. Statistical significance is indicated by asterisks: ****~($p<0.0001$), ***~($p<0.001$), and **~($p<0.01$). }
\label{fig1}
\end{figure}

\begin{figure}[htbp]
\centering
\includegraphics[width=1\textwidth]{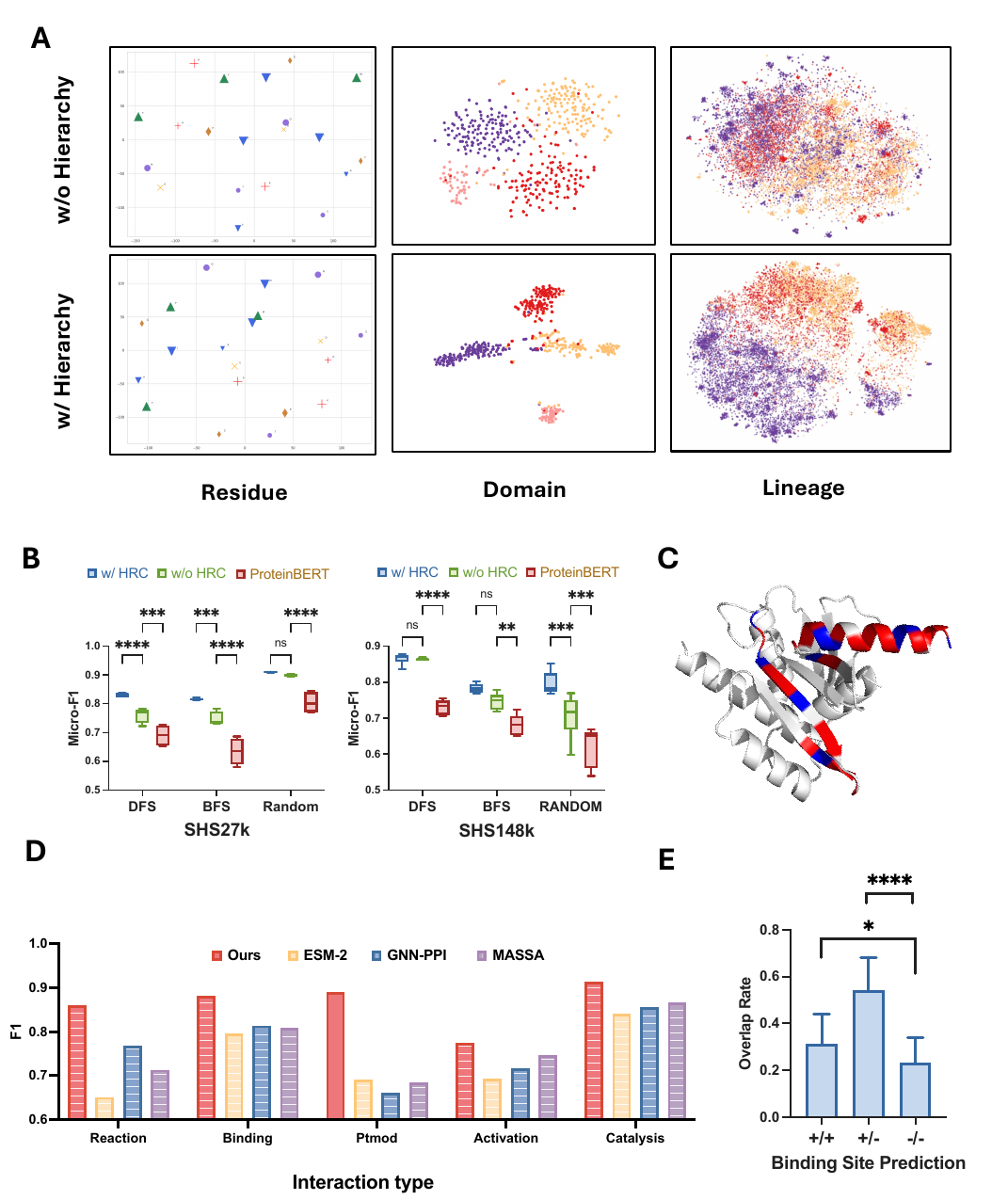}

\label{fig2}
\end{figure}

\begin{figure}[htbp]
    \caption{Protein embeddings and attention analysis for PPI prediction. 
(A)~The left column highlights residue-level embeddings colored by amino acid biophysical properties. The middle column annotates embeddings according to protein domain classifications. The right column distinguishes proteins by lineage or organism classifications. The comparative visualization clearly illustrates improved clustering and distinct separation patterns in the lower row (w/ Hierarchy). t-SNE projections visualize information extracted by protein representation models (upper row: without hierarchical pre-training on Swiss-Prot, and lower row: with hierarchical pre-training on Swiss-Prot). Residue:~Visualization of residue embedding. Two-dimensional projection of amino acid token embeddings colored and shaped by physicochemical properties (e.g., polarity, hydrophobicity, and size). The separation among groups reflects the model’s ability to learn property-aware representations. Domain:~Visualization of 4 different domain embedding(~Protein kinase domain,~WD40 repeats,~Immunoglobulin domain,~Ankyrin repeat). Lineage:~Visualization of three different lineage embeddings (Actinobacteria, Bacteroidetes, and Firmicutes) on the Beta-lactamase family PF00144 from Pfam.  (C)~Identification of binding site(PDB id:~3VZ9). The area in red represents the predicted binding site region, while the blue area represents the missed area of binding sites. The white area depicts the non-essential area of the protein. (D)Distributions of AUPR scores on 5 PPI types. Bar plots showing F1 scores of four models (Ours, ESM-2, GNN-PPI, and MASSA) across five distinct interaction types (Reaction, Binding, PTMod, Activation, and Catalysis) under DFS sampling strategies on the SHS27k dataset. (E)~Bar plots display the overlap rates for binding site prediction across 32 protein complexes under three model configurations: hierarchical information combined with flat annotations ($+/+$), hierarchical information without flat annotations ($+/-$), and neither hierarchical nor flat annotations ($-/-$). }
    \label{fig:my_figure}
\end{figure}

\subsection{Modeling Hierarchical Relationships Among Proteins Improves Performance}\label{subsec22}

As shown in Fig.~2b and 3b, the removal of hierarchical relationships among protein properties resulted in a marked decrease in predictive accuracy across all intra-species datasets. The F1 scores dropped by up to 15\%\ when hierarchical labels were omitted, underscoring the significant contribution of structured biological information. Notably, the performance degradation was most pronounced on “hard” test pairs, where both interacting proteins were unseen during training, suggesting that hierarchical modeling is particularly beneficial for generalizing to novel or under-annotated proteins.

Further analysis revealed that integrating hierarchical relationships facilitates the learning of protein representations that more effectively capture functional similarities and evolutionary relationships. This structural prior enables the model to cluster proteins with shared functional characteristics more closely in the embedding space, thereby improving the discrimination between interacting and non-interacting pairs.

To evaluate whether hierarchical supervision helps the model capture biologically meaningful structure, we visualized the learned protein embeddings at three levels-residue, domain, and lineage—with and without hierarchical constraints, as shown in Fig.3A. In the absence of hierarchical supervision, the resulting representations show limited organization, with protein points dispersed and clusters poorly defined across all annotation levels. When hierarchical relationships are incorporated, clear and compact clustering emerges at the domain and lineage levels, indicating that the model is able to learn functionally relevant structure by leveraging hierarchical information.

Furthermore, we examined the specific contribution of non-hierarchical information by excluding non-hierarchical annotations from the model. As shown in Fig.~2D, removing non-hierarchical information and relying solely on hierarchical annotations led to a noticeable decrease in intra-species prediction accuracy. Visualizations of the learned embeddings at both residue and lineage levels reveal that, with only family supervision, the representations remain scattered and lack clear clustering. Interestingly, as shown in Fig.~3E, this ablation also resulted in improved performance in binding site prediction. Therefore, while this reduction indicates that non-hierarchical annotations contain valuable information for general prediction tasks, the improved binding site predictions suggest hierarchical annotations alone effectively highlight more precise functional interactions.

Collectively, these results demonstrate that modeling hierarchical relationships among protein properties is critical for enhancing predictive performance, especially in challenging prediction scenarios involving unfamiliar proteins.

\subsection{HIPPO Extends to Cross-Species PPI Prediction with Leading Performance}\label{subsec23}

Cross-species protein-protein interaction (PPI) prediction addresses the challenge of inferring interactions in less-studied species by leveraging knowledge learned from well-studied organisms, such as humans. Unlike traditional approaches that rely on species-specific training data, our framework transfers predictive power across species boundaries, enabling PPI prediction where experimental data are scarce or unavailable. In this section, we (1) systematically evaluate the model’s ability to predict PPIs in six species after training on human datasets-specifically, Escherichia coli, Saccharomyces cerevisiae (yeast), Mus musculus (mouse), Caenorhabditis elegans, Arabidopsis thaliana, and Drosophila melanogaster, all of which represent widely studied model organisms in genetics, developmental biology, and plant science; (2) benchmark against existing methods; and (3) analyze the factors underlying successful cross-species generalization.

As summarized in Fig.~4, our model consistently outperforms existing methods, including GNN-PPI, ProteinBERT, ESM-2, PIPR, and INTREPPID, across all cross-species datasets. The proposed approach achieves the highest F1 scores in each species, demonstrating its robustness and superior predictive power in challenging cross-species settings. These results highlight the capacity of our hierarchical, multimodal framework to learn transferable protein representations that generalize well to novel proteomes.
The performance suggests that the model effectively captures conserved functional features and interaction patterns shared among diverse organisms. Notably, the model’s robust results in intra-species PPI prediction provide a solid foundation for its success in cross-species scenarios, as both settings benefit from the hierarchical and transferable representations learned by HIPPO. This broad applicability is particularly valuable for predicting PPIs in non-model species, where experimental data are limited. Overall, these findings underscore the potential of our approach for large-scale, cross-species interactome mapping.

 \subsection{HIPPO Identifies Functional Binding Sites in Interaction Complexes}\label{subsec24}

To investigate the underlying reasons for the improved prediction accuracy achieved by incorporating hierarchical information, we analyzed HIPPO's internal attention mechanism. Specifically, we utilized attention weights extracted from the model to identify functional three-dimensional binding site positions within protein–protein interaction (PPI) complexes.
Fig.~3C provides a structural visualization based on the crystal structure of the chicken Spc24-Spc25 globular domain (PDB ID: 3vz9). Regions shown in red correspond to accurately predicted binding sites, while those in blue indicate missed binding sites. Notably, incorporating hierarchical contrastive learning results in a high overlap rate of 0.71875 for binding site prediction in this complex, compared to 0.40625 without this hierarchical supervision, highlighting HIPPO’s enhanced capability to precisely identify interaction interfaces.

In Fig.~3E, we further evaluated binding site prediction across a broader set of 35 protein complexes, comparing three model configurations: hierarchical information combined with flat annotations (+/+), hierarchical information without flat annotations (+/–), and neither hierarchical nor flat annotations (–/–). Models incorporating hierarchical annotations (+/+ and +/–) significantly outperform the annotation-free model (–/–), underscoring the value of hierarchical supervision. Interestingly, the addition of flat annotations alongside hierarchical annotations (+/+) slightly decreases prediction performance compared to the hierarchical-only scenario (+/–). These results indicate that while hierarchical information is crucial for identifying interaction sites, the introduction of non-hierarchical annotations may diminish the model’s ability to precisely pinpoint functional binding regions.

Collectively, this analysis reveals that HIPPO’s hierarchical attention mechanism effectively captures biologically meaningful binding sites, contributing significantly to the overall accuracy of PPI predictions.

\begin{figure}[h]
\centering
\includegraphics[width=1\textwidth]{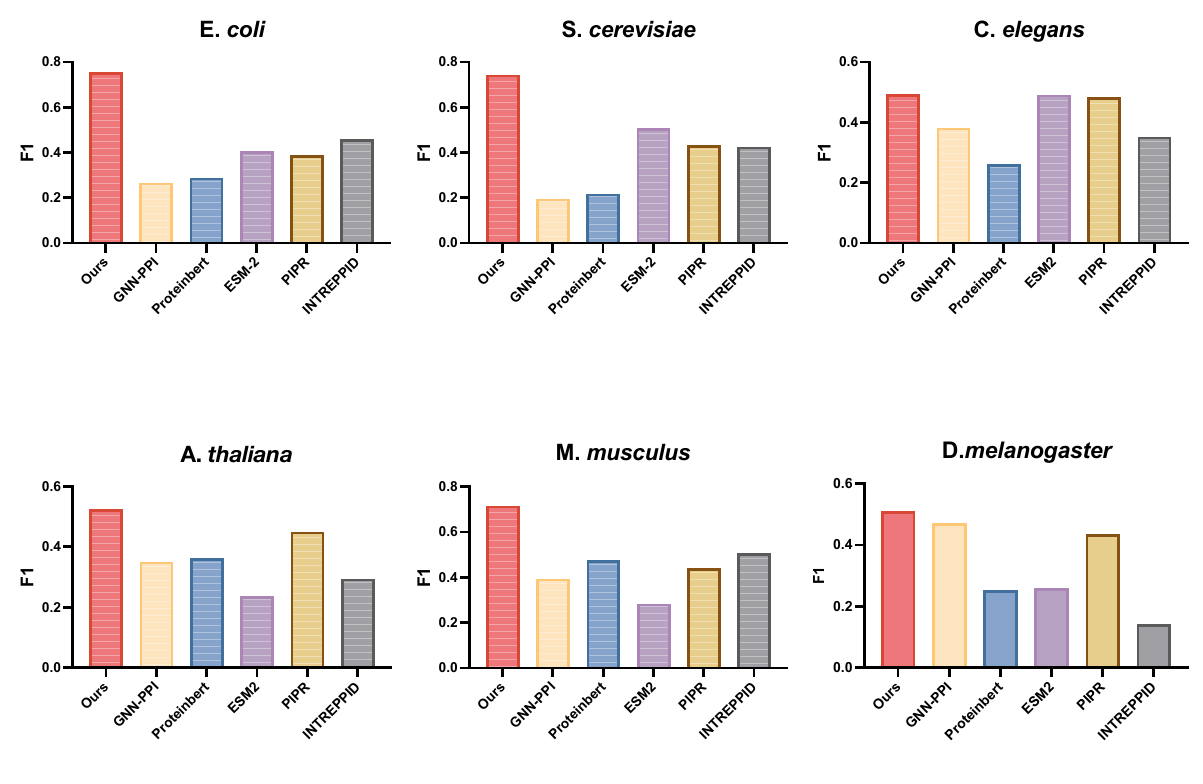}
\caption{Performance comparison of models in predicting cross-species protein–protein interactions (PPIs). Bar plots depict Micro-F1 scores across five cross-species datasets: Escherichia coli (E.coli), Saccharomyces cerevisiae (S.cerevisiae), Caenorhabditis elegans (C.elegans), Mus musculus (M.musculus),Arabidopsis thaliana (A.thaliana), and Drosophila melanogaster(D.melanogaster). HIPPO(Ours) is benchmarked against five state-of-the-art models: GNN-PPI, ProteinBERT, ESM-2, PIPR, and INTREPPID. }
\label{fig2}
\end{figure}

\section{Methods}\label{sec20}

\subsection{Data Preparation and Analysis }\label{subsec23}

During the pretraining stage, we employed the Swiss-Prot dataset from UniProtKB \cite{bairoch2000swiss} due to its extensive protein data coverage and high-quality, manually curated annotations. Protein attributes include various protein families and clans accessible through the Pfam database \cite{mistry2021pfam}, and annotations curated within the UniProtKB Keywords section. The keywords section incorporates controlled vocabulary terms manually annotated to include Gene Ontology (GO) terms, disease associations, protein domains, ligands, and post-translational modifications (PTMs).

During protein–protein interaction (PPI) prediction stage, we utilized three datasets derived from STRING: the full STRING dataset, SHS148k, and SHS27k. The STRING dataset contains 1,048,575 interactions among 16,073 proteins. SHS27k and SHS148k were created by selecting proteins longer than 50 amino acids with less than 40\%\ sequence identity to form more challenging subsets. SHS27k comprises 63,408 interactions among 1,690 proteins, while SHS148k includes 36,902 interactions among 5,189 proteins. Three partition methods—Random, Breadth-First Search (BFS), and Depth-First Search (DFS)—were employed for dataset splitting \cite{lv2021learning}. To rigorously assess model generalization, we evaluated performance on unfamiliar proteins within the test set by checking if interacting proteins existed in the training set, with detailed splitting procedures described in the Results section.

Hierarchical protein labels represent multiple interconnected annotations related to protein functions and residue-level details, organized in a tree structure. In this hierarchy, leaf nodes correspond to unique sequence identifiers, while non-leaf nodes indicate evolutionary classifications at various hierarchical levels. Higher hierarchy levels (e.g., \textit{Clan} $l_1$) represent broader evolutionary relationships, situated closer to the root, whereas lower levels (e.g., \textit{Family} $l_2$) represent narrower classifications. Clans integrate multiple related protein domain families, reflecting extensive evolutionary connections and similarities. Positive sequence pairs at a given hierarchical level $l \in L$ are defined as sequences sharing common ancestry up to level $l$ but diverging thereafter. As illustrated in Fig.~1, a pair at the clan level $l_1$ implies their lowest common ancestor is at this hierarchical level. Our framework comprises 6,329 distinct families and 621 clans, collectively forming a hierarchical clan-family tree capturing intrinsic protein properties such as evolutionary relationships, sequence similarities, and structural homologies \cite{paysan2025pfam}.

\subsection{Construction of network graph for PPI complexes}

Given a protein set \( \mathcal{P} = \{p_0, p_1, \dots, p_n\} \) and a corresponding set of protein–protein interactions (PPIs)
\[
\mathcal{X} = \{x_{ij} = \{p_i, p_j\} \mid i \neq j, p_i, p_j \in \mathcal{P} \},
\]
we define the PPI label space as
\[
\mathcal{L} = \{l_0, l_1, \dots, l_n\},
\]
where each interaction \( x_{ij} \) is associated with a label set \( y_{ij} \subseteq \mathcal{L} \), indicating multiple potential interaction types. The full dataset is denoted as
\[
\mathcal{D} = \{(x_{ij}, y_{ij}) \mid x_{ij} \in \mathcal{X} \},
\]
and can be naturally represented as a graph \( \mathcal{G} = (\mathcal{P}, \mathcal{X}) \), where proteins are nodes and interactions are labeled edges.

To incorporate functional and evolutionary context, we encode each protein into a hierarchical embedding informed by family and clan-level annotations. These representations capture evolutionary lineage and structural similarity, which are essential for understanding protein interaction mechanisms.

We formulate the multi-label PPI task as learning a prediction function
\[
\mathcal{F}: x_{ij} \rightarrow \hat{y}_{ij},
\]
using a training subset \( \mathcal{X}_{\text{train}} \subseteq \mathcal{X} \), with evaluation performed on a disjoint test set \( \mathcal{X}_{\text{test}} \), satisfying \( \mathcal{X}_{\text{train}} \cup \mathcal{X}_{\text{test}} = \mathcal{X} \).

To implement this, we adopt the Graph Isomorphism Network (GIN)~\cite{xu2018powerful} as the backbone encoder. GIN aggregates neighborhood information from each protein node to produce discriminative embeddings. For each protein pair \( x_{ij} \), the embeddings \( g_{p_i} \) and \( g_{p_j} \) are combined using a dot product, followed by a fully connected (FC) layer for final prediction:
\[
\hat{y}_{ij} = \text{FC}(g_{p_i} \cdot g_{p_j}).
\]
The model is trained using a binary cross-entropy loss over all interaction types:
\[
L = \sum_{k=0}^{n} \sum_{x_{ij} \in \mathcal{X}_{\text{train}}} \left( -y_{ij}^{k} \log \hat{y}_{ij}^{k} - \left( 1 - y_{ij}^{k} \right) \log \left( 1 - \hat{y}_{ij}^{k} \right) \right),
\]
where \( y_{ij}^{k} \) and \( \hat{y}_{ij}^{k} \) denote the ground truth and predicted probability of the \( k \)-th interaction type for the protein pair \( x_{ij} \).

\subsection{Protein Representation Learning with Functional and Hierarchical Supervision}\label{subsec:prot_rep_learn}

To enable effective downstream prediction of protein–protein interactions, we design a pre-training stage that integrates both hierarchical and functional annotation sources into protein sequence representations. Our approach consists of two complementary components: hierarchical relationship contrastive learning and multimodal sequence-annotation alignment.

\paragraph{Hierarchical Relationship Contrastive Learning.}
To reflect the biological reality that protein function and structure are organized hierarchically (e.g., clans, families, domains), we explicitly preserve this hierarchy during pre-training. Drawing on the Hierarchical Multi-label Constraint Enforcing Contrastive Loss (HiMulConE)~\cite{zhang2022use}, we encourage the model to maintain consistent representations across hierarchical levels.  
Given a set of hierarchical label levels $L$, the constraint ensures that confidence in higher-level ancestors (e.g., clan) is never less than that of their lower-level descendants (e.g., family). The hierarchical contrastive loss is defined as:
\begin{equation}
\mathcal{L}_{\text{HC}} = \sum_{l \in L} \frac{1}{|L|} \sum_{i \in I} \frac{-\lambda_l}{|P(i)|} \sum_{p_l \in P_l} \max\left(L^{\text{pair}}(i, p^i_l), L^{\text{pair}}_{\text{max}}(l-1)\right),
\end{equation}
where $L^{\text{pair}}_{\text{max}}(l)$ is the maximum loss among all positive pairs at level $l$, and $L^{\text{pair}}(i, p^i_l)$ is given by:
\begin{equation}
L^{\text{pair}}(i, p^i_l) = \log \frac{\exp(f_i \cdot f^{l}_{p} / \tau)}{\sum_{a \in A \setminus i} \exp(f_i \cdot f_a / \tau)},
\end{equation}
with $f_i$ and $f_p^l$ denoting protein feature representations at level $l$, and $\lambda_l$ being a weighting factor for level $l$.

\paragraph{Multimodal Sequence-Annotation Alignment.}
We next aim to align protein sequence embeddings with their corresponding functional annotation embeddings, thereby injecting curated biological knowledge into the learned representations. Specifically, we adapt the sequence and annotation encoding modules from ProteinBERT~\cite{brandes2022proteinbert} and perform end-to-end pre-training.  
Protein sequences are encoded using a protein language model (PLM), producing local sequence representations $\{\mathbf{z}^S_i\}_{i=1}^N$ for a batch of $N$ proteins. In parallel, an annotation language model (ALM) generates functional context embeddings $\{\mathbf{z}^T_i\}_{i=1}^N$ from textual descriptions of protein properties, families, and domains~\cite{jin2019probing}.  
To maximize the agreement between matching sequence–annotation pairs while discouraging spurious alignments, we employ a symmetric InfoNCE loss~\cite{journals/corr/abs-1807-03748}:
\begin{equation}
\mathcal{L}_{\text{SAC}} = -\frac{1}{2N} \sum_{i=1}^{N} \left(
    \log \frac{\exp(\mathbf{z}_i^S \cdot \mathbf{z}_i^A / \tau)}{\sum_{j=1}^{N} \exp(\mathbf{z}_i^S \cdot \mathbf{z}_j^T / \tau)} + 
    \log \frac{\exp(\mathbf{z}_i^A \cdot \mathbf{z}_i^T / \tau)}{\sum_{j=1}^{N} \exp(\mathbf{z}_j^A \cdot \mathbf{z}_i^T / \tau)}
\right),
\end{equation}
where $\tau$ is a temperature parameter.  
To further distinguish between truly corresponding (positive) and mismatched (negative) sequence–annotation pairs, we introduce a Sequence-Annotation Matching (SAM) loss using focal loss:
\begin{equation}
\mathcal{L}_{\text{SAM}} = - \frac{1}{N} \sum_{i=1}^{N} \Big( 
    \alpha (1 - p_i)^\gamma \log(p_i) \cdot y^{\mathrm{SAM}}_i + 
    (1 - \alpha) p_i^\gamma \log(1 - p_i) \cdot (1 - y^{\mathrm{SAM}}_i)
\Big),
\end{equation}
where $p_i$ denotes the predicted probability that the $i$-th sequence–annotation pair is a true match, $y^{\mathrm{SAM}}_i \in \{0, 1\}$ is the binary indicator for the $i$-th pair (with $y^{\mathrm{SAM}}_i = 1$ for positive pairs and $0$ otherwise), $\alpha$ is a class balancing parameter, and $\gamma$ is the focusing parameter.

\paragraph{Overall Pre-training Objective.}
The total pre-training loss is a weighted sum of the above objectives:
\begin{equation}
    \min_\theta \left[ \mathcal{L}_{\text{HC}} + \mathcal{L}_{\text{SAC}} + \mathcal{L}_{\text{SAM}} \right],
\end{equation}
where $\theta$ includes all trainable parameters of the PLM, ALM, and projection heads. 
\subsection{Model training details}\label{subsec29}
Given the significant imbalance in PPI types within the datasets, micro-F1 is preferred over macro-F1 as an evaluation metric for multi-label PPI type prediction. We divide the PPIs into training(80\%), and validation(20\%) set across all baselines, as reported in previous studies\cite{lv2021learning,gao2023hierarchical}. We then select the model that achieves the best performance on the validation set to evaluate the micro-F1 scores on the test data. Additionally, each set of experiments is run five times with different random seeds, and we report the best performance.

We compare our model against several established protein representation learning methods, including GNN-PPI~\cite{lv2021learning}. In addition, we consider language models pretrained on large-scale sequence datasets, such as ESM-2~\cite{rives2019biological}. For cross-species prediction, we also include INTREPPID~\cite{intrepppid}, a model designed for cross-species PPI prediction.

\section{Discussion}\label{sec21}

In this study, we introduced the HIPPO framework to systematically explore how hierarchical representations of protein function and domain annotation can advance protein–protein interaction (PPI) prediction across both intra- and cross-species contexts. By assembling a comprehensive dataset that aligns protein sequences with a diverse set of functional annotations-including domain/family information, Gene Ontology terms, and post-translational modifications (PTMs)-we enable fine-grained analysis of the effects of hierarchical knowledge integration.

Experimental results show that HIPPO delivers high predictive accuracy for PPIs both within individual species and across species boundaries, demonstrating its potential for generalizing to less-studied organisms. Notably, the model excels in challenging cases involving binding events. Attention-based analyses indicate that HIPPO learns biochemically meaningful patterns at the residue level, such as consistent attention distributions among residues with similar side chains. Nevertheless, limitations remain in modeling more complex and nuanced relationships between proteins, due in part to the coverage and diversity of current annotation sources. Future work will focus on expanding the dataset-incorporating site-specific PTMs and structural features-and developing more sophisticated approaches for modeling protein functional contexts and interactions.

\section{Conclusion}\label{sec22}

In summary, we present a novel computational framework that explicitly incorporates hierarchical biological relationships into protein representation learning, enabling robust and generalizable PPI prediction in both intra-species and cross-species settings. By alleviating the dependence on experimentally resolved structural data, HIPPO facilitates accurate and scalable PPI prediction, even for species where structural information is scarce or unavailable. This makes our approach particularly well-suited for guiding biological discovery and candidate prioritization in less-characterized or rare species.

Moreover, our results demonstrate that the learned representations capture meaningful biological features, as evidenced by interpretable patterns in binding site and PTM analyses across species. Looking forward, we aim to address the current model’s limitations by extending the representation of complex and dynamic protein relationships, including context-specific interactions and multi-protein complexes, to further advance the capabilities of computational interactomics.


\begin{thebibliography}{27}
\bibliographystyle{sn-nature}
\ifx \bisbn   \undefined \def \bisbn  #1{ISBN #1}\fi
\ifx \binits  \undefined \def \binits#1{#1}\fi
\ifx \bauthor  \undefined \def \bauthor#1{#1}\fi
\ifx \batitle  \undefined \def \batitle#1{#1}\fi
\ifx \bjtitle  \undefined \def \bjtitle#1{#1}\fi
\ifx \bvolume  \undefined \def \bvolume#1{\textbf{#1}}\fi
\ifx \byear  \undefined \def \byear#1{#1}\fi
\ifx \bissue  \undefined \def \bissue#1{#1}\fi
\ifx \bfpage  \undefined \def \bfpage#1{#1}\fi
\ifx \blpage  \undefined \def \blpage #1{#1}\fi
\ifx \burl  \undefined \def \burl#1{\textsf{#1}}\fi
\ifx \doiurl  \undefined \def \doiurl#1{\url{https://doi.org/#1}}\fi
\ifx \betal  \undefined \def \betal{\textit{et al.}}\fi
\ifx \binstitute  \undefined \def \binstitute#1{#1}\fi
\ifx \binstitutionaled  \undefined \def \binstitutionaled#1{#1}\fi
\ifx \bctitle  \undefined \def \bctitle#1{#1}\fi
\ifx \beditor  \undefined \def \beditor#1{#1}\fi
\ifx \bpublisher  \undefined \def \bpublisher#1{#1}\fi
\ifx \bbtitle  \undefined \def \bbtitle#1{#1}\fi
\ifx \bedition  \undefined \def \bedition#1{#1}\fi
\ifx \bseriesno  \undefined \def \bseriesno#1{#1}\fi
\ifx \blocation  \undefined \def \blocation#1{#1}\fi
\ifx \bsertitle  \undefined \def \bsertitle#1{#1}\fi
\ifx \bsnm \undefined \def \bsnm#1{#1}\fi
\ifx \bsuffix \undefined \def \bsuffix#1{#1}\fi
\ifx \bparticle \undefined \def \bparticle#1{#1}\fi
\ifx \barticle \undefined \def \barticle#1{#1}\fi
\bibcommenthead
\ifx \bconfdate \undefined \def \bconfdate #1{#1}\fi
\ifx \botherref \undefined \def \botherref #1{#1}\fi
\ifx \url \undefined \def \url#1{\textsf{#1}}\fi
\ifx \bchapter \undefined \def \bchapter#1{#1}\fi
\ifx \bbook \undefined \def \bbook#1{#1}\fi
\ifx \bcomment \undefined \def \bcomment#1{#1}\fi
\ifx \oauthor \undefined \def \oauthor#1{#1}\fi
\ifx \citeauthoryear \undefined \def \citeauthoryear#1{#1}\fi
\ifx \endbibitem  \undefined \def \endbibitem {}\fi
\ifx \bconflocation  \undefined \def \bconflocation#1{#1}\fi
\ifx \arxivurl  \undefined \def \arxivurl#1{\textsf{#1}}\fi
\csname PreBibitemsHook\endcsname

\bibitem[\protect\citeauthoryear{Br{\"u}ckner et~al.}{2009}]{bruckner2009yeast}
\begin{barticle}
\bauthor{\bsnm{Br{\"u}ckner}, \binits{A.}},
\bauthor{\bsnm{Polge}, \binits{C.}},
\bauthor{\bsnm{Lentze}, \binits{N.}},
\bauthor{\bsnm{Auerbach}, \binits{D.}},
\bauthor{\bsnm{Schlattner}, \binits{U.}}:
\batitle{Yeast two-hybrid, a powerful tool for systems biology}.
\bjtitle{International journal of molecular sciences}
\bvolume{10}(\bissue{6}),
\bfpage{2763}--\blpage{2788}
(\byear{2009})
\end{barticle}
\endbibitem

\bibitem[\protect\citeauthoryear{Kaboord and Perr}{2008}]{kaboord2008isolation}
\begin{botherref}
\oauthor{\bsnm{Kaboord}, \binits{B.}},
\oauthor{\bsnm{Perr}, \binits{M.}}:
Isolation of proteins and protein complexes by immunoprecipitation.
2D PAGE: sample preparation and fractionation,
349--364
(2008)
\end{botherref}
\endbibitem

\bibitem[\protect\citeauthoryear{Aronheim et~al.}{1997}]{aronheim1997isolation}
\begin{barticle}
\bauthor{\bsnm{Aronheim}, \binits{A.}},
\bauthor{\bsnm{Zandi}, \binits{E.}},
\bauthor{\bsnm{Hennemann}, \binits{H.}},
\bauthor{\bsnm{Elledge}, \binits{S.J.}},
\bauthor{\bsnm{Karin}, \binits{M.}}:
\batitle{Isolation of an ap-1 repressor by a novel method for detecting
  protein-protein interactions}.
\bjtitle{Molecular and cellular biology}
\bvolume{17}(\bissue{6}),
\bfpage{3094}--\blpage{3102}
(\byear{1997})
\end{barticle}
\endbibitem

\bibitem[\protect\citeauthoryear{Tan et~al.}{2016}]{tan2016trifunctional}
\begin{barticle}
\bauthor{\bsnm{Tan}, \binits{D.}},
\bauthor{\bsnm{Li}, \binits{Q.}},
\bauthor{\bsnm{Zhang}, \binits{M.-J.}},
\bauthor{\bsnm{Liu}, \binits{C.}},
\bauthor{\bsnm{Ma}, \binits{C.}},
\bauthor{\bsnm{Zhang}, \binits{P.}},
\bauthor{\bsnm{Ding}, \binits{Y.-H.}},
\bauthor{\bsnm{Fan}, \binits{S.-B.}},
\bauthor{\bsnm{Tao}, \binits{L.}},
\bauthor{\bsnm{Yang}, \binits{B.}}, \betal:
\batitle{Trifunctional cross-linker for mapping protein-protein interaction
  networks and comparing protein conformational states}.
\bjtitle{Elife}
\bvolume{5},
\bfpage{12509}
(\byear{2016})
\end{barticle}
\endbibitem

\bibitem[\protect\citeauthoryear{Cho et~al.}{2020}]{cho2020proximity}
\begin{barticle}
\bauthor{\bsnm{Cho}, \binits{K.F.}},
\bauthor{\bsnm{Branon}, \binits{T.C.}},
\bauthor{\bsnm{Udeshi}, \binits{N.D.}},
\bauthor{\bsnm{Myers}, \binits{S.A.}},
\bauthor{\bsnm{Carr}, \binits{S.A.}},
\bauthor{\bsnm{Ting}, \binits{A.Y.}}:
\batitle{Proximity labeling in mammalian cells with turboid and split-turboid}.
\bjtitle{Nature Protocols}
\bvolume{15}(\bissue{12}),
\bfpage{3971}--\blpage{3999}
(\byear{2020})
\end{barticle}
\endbibitem

\bibitem[\protect\citeauthoryear{Singh et~al.}{2023}]{singh2023contrastive}
\begin{barticle}
\bauthor{\bsnm{Singh}, \binits{R.}},
\bauthor{\bsnm{Sledzieski}, \binits{S.}},
\bauthor{\bsnm{Bryson}, \binits{B.}},
\bauthor{\bsnm{Cowen}, \binits{L.}},
\bauthor{\bsnm{Berger}, \binits{B.}}:
\batitle{Contrastive learning in protein language space predicts interactions
  between drugs and protein targets}.
\bjtitle{Proceedings of the National Academy of Sciences}
\bvolume{120}(\bissue{24}),
\bfpage{2220778120}
(\byear{2023})
\end{barticle}
\endbibitem

\bibitem[\protect\citeauthoryear{Zhang et~al.}{2022}]{zhang2022application}
\begin{barticle}
\bauthor{\bsnm{Zhang}, \binits{Y.}},
\bauthor{\bsnm{Luo}, \binits{M.}},
\bauthor{\bsnm{Wu}, \binits{P.}},
\bauthor{\bsnm{Wu}, \binits{S.}},
\bauthor{\bsnm{Lee}, \binits{T.-Y.}},
\bauthor{\bsnm{Bai}, \binits{C.}}:
\batitle{Application of computational biology and artificial intelligence in
  drug design}.
\bjtitle{International journal of molecular sciences}
\bvolume{23}(\bissue{21}),
\bfpage{13568}
(\byear{2022})
\end{barticle}
\endbibitem

\bibitem[\protect\citeauthoryear{Jumper et~al.}{2021}]{jumper2021highly}
\begin{barticle}
\bauthor{\bsnm{Jumper}, \binits{J.}},
\bauthor{\bsnm{Evans}, \binits{R.}},
\bauthor{\bsnm{Pritzel}, \binits{A.}},
\bauthor{\bsnm{Green}, \binits{T.}},
\bauthor{\bsnm{Figurnov}, \binits{M.}},
\bauthor{\bsnm{Ronneberger}, \binits{O.}},
\bauthor{\bsnm{Tunyasuvunakool}, \binits{K.}},
\bauthor{\bsnm{Bates}, \binits{R.}},
\bauthor{\bsnm{{\v{Z}}{\'\i}dek}, \binits{A.}},
\bauthor{\bsnm{Potapenko}, \binits{A.}}, \betal:
\batitle{Highly accurate protein structure prediction with alphafold}.
\bjtitle{nature}
\bvolume{596}(\bissue{7873}),
\bfpage{583}--\blpage{589}
(\byear{2021})
\end{barticle}
\endbibitem

\bibitem[\protect\citeauthoryear{Kim et~al.}{2023}]{kim2023functional}
\begin{barticle}
\bauthor{\bsnm{Kim}, \binits{G.B.}},
\bauthor{\bsnm{Kim}, \binits{J.Y.}},
\bauthor{\bsnm{Lee}, \binits{J.A.}},
\bauthor{\bsnm{Norsigian}, \binits{C.J.}},
\bauthor{\bsnm{Palsson}, \binits{B.O.}},
\bauthor{\bsnm{Lee}, \binits{S.Y.}}:
\batitle{Functional annotation of enzyme-encoding genes using deep learning
  with transformer layers}.
\bjtitle{Nature Communications}
\bvolume{14}(\bissue{1}),
\bfpage{7370}
(\byear{2023})
\end{barticle}
\endbibitem

\bibitem[\protect\citeauthoryear{Lv et~al.}{2021}]{lv2021learning}
\begin{botherref}
\oauthor{\bsnm{Lv}, \binits{G.}},
\oauthor{\bsnm{Hu}, \binits{Z.}},
\oauthor{\bsnm{Bi}, \binits{Y.}},
\oauthor{\bsnm{Zhang}, \binits{S.}}:
Learning unknown from correlations: graph neural network for
  inter-novel-protein interaction prediction.
arXiv preprint arXiv:2105.06709
(2021)
\end{botherref}
\endbibitem

\bibitem[\protect\citeauthoryear{Radford et~al.}{2021}]{radford2021learning}
\begin{bchapter}
\bauthor{\bsnm{Radford}, \binits{A.}},
\bauthor{\bsnm{Kim}, \binits{J.W.}},
\bauthor{\bsnm{Hallacy}, \binits{C.}},
\bauthor{\bsnm{Ramesh}, \binits{A.}},
\bauthor{\bsnm{Goh}, \binits{G.}},
\bauthor{\bsnm{Agarwal}, \binits{S.}},
\bauthor{\bsnm{Sastry}, \binits{G.}},
\bauthor{\bsnm{Askell}, \binits{A.}},
\bauthor{\bsnm{Mishkin}, \binits{P.}},
\bauthor{\bsnm{Clark}, \binits{J.}}, \betal:
\bctitle{Learning transferable visual models from natural language
  supervision}.
In: \bbtitle{International Conference on Machine Learning},
pp. \bfpage{8748}--\blpage{8763}
(\byear{2021}).
\bcomment{PMLR}
\end{bchapter}
\endbibitem

\bibitem[\protect\citeauthoryear{Oord et~al.}{2018}]{oord2018representation}
\begin{botherref}
\oauthor{\bsnm{Oord}, \binits{A.v.d.}},
\oauthor{\bsnm{Li}, \binits{Y.}},
\oauthor{\bsnm{Vinyals}, \binits{O.}}:
Representation learning with contrastive predictive coding.
arXiv preprint arXiv:1807.03748
(2018)
\end{botherref}
\endbibitem

\bibitem[\protect\citeauthoryear{Ramesh et~al.}{2022}]{ramesh2022hierarchical}
\begin{barticle}
\bauthor{\bsnm{Ramesh}, \binits{A.}},
\bauthor{\bsnm{Dhariwal}, \binits{P.}},
\bauthor{\bsnm{Nichol}, \binits{A.}},
\bauthor{\bsnm{Chu}, \binits{C.}},
\bauthor{\bsnm{Chen}, \binits{M.}}:
\batitle{Hierarchical text-conditional image generation with clip latents}.
\bjtitle{arXiv preprint arXiv:2204.06125}
\bvolume{1}(\bissue{2}),
\bfpage{3}
(\byear{2022})
\end{barticle}
\endbibitem

\bibitem[\protect\citeauthoryear{Ashburner et~al.}{2000}]{ashburner2000gene}
\begin{barticle}
\bauthor{\bsnm{Ashburner}, \binits{M.}},
\bauthor{\bsnm{Ball}, \binits{C.A.}},
\bauthor{\bsnm{Blake}, \binits{J.A.}},
\bauthor{\bsnm{Botstein}, \binits{D.}},
\bauthor{\bsnm{Butler}, \binits{H.}},
\bauthor{\bsnm{Cherry}, \binits{J.M.}},
\bauthor{\bsnm{Davis}, \binits{A.P.}},
\bauthor{\bsnm{Dolinski}, \binits{K.}},
\bauthor{\bsnm{Dwight}, \binits{S.S.}},
\bauthor{\bsnm{Eppig}, \binits{J.T.}}, \betal:
\batitle{Gene ontology: tool for the unification of biology}.
\bjtitle{Nature genetics}
\bvolume{25}(\bissue{1}),
\bfpage{25}--\blpage{29}
(\byear{2000})
\end{barticle}
\endbibitem

\bibitem[\protect\citeauthoryear{Wright and Dyson}{2015}]{Wright2015IDP}
\begin{barticle}
\bauthor{\bsnm{Wright}, \binits{P.E.}},
\bauthor{\bsnm{Dyson}, \binits{H.J.}}:
\batitle{Intrinsically disordered proteins in cellular signalling and
  regulation}.
\bjtitle{Nature Reviews Molecular Cell Biology}
\bvolume{16}(\bissue{1}),
\bfpage{18}--\blpage{29}
(\byear{2015})
\doiurl{10.1038/nrm3920}
\end{barticle}
\endbibitem

\bibitem[\protect\citeauthoryear{Gao et~al.}{2023}]{gao2023hierarchical}
\begin{barticle}
\bauthor{\bsnm{Gao}, \binits{Z.}},
\bauthor{\bsnm{Jiang}, \binits{C.}},
\bauthor{\bsnm{Zhang}, \binits{J.}},
\bauthor{\bsnm{Jiang}, \binits{X.}},
\bauthor{\bsnm{Li}, \binits{L.}},
\bauthor{\bsnm{Zhao}, \binits{P.}},
\bauthor{\bsnm{Yang}, \binits{H.}},
\bauthor{\bsnm{Huang}, \binits{Y.}},
\bauthor{\bsnm{Li}, \binits{J.}}:
\batitle{Hierarchical graph learning for protein--protein interaction}.
\bjtitle{Nature Communications}
\bvolume{14}(\bissue{1}),
\bfpage{1093}
(\byear{2023})
\end{barticle}
\endbibitem

\bibitem[\protect\citeauthoryear{Wu et~al.}{2024}]{wu2024mape}
\begin{botherref}
\oauthor{\bsnm{Wu}, \binits{L.}},
\oauthor{\bsnm{Tian}, \binits{Y.}},
\oauthor{\bsnm{Huang}, \binits{Y.}},
\oauthor{\bsnm{Li}, \binits{S.}},
\oauthor{\bsnm{Lin}, \binits{H.}},
\oauthor{\bsnm{Chawla}, \binits{N.V.}},
\oauthor{\bsnm{Li}, \binits{S.Z.}}:
Mape-ppi: Towards effective and efficient protein-protein interaction
  prediction via microenvironment-aware protein embedding.
arXiv preprint arXiv:2402.14391
(2024)
\end{botherref}
\endbibitem

\bibitem[\protect\citeauthoryear{Bairoch and Apweiler}{2000}]{bairoch2000swiss}
\begin{barticle}
\bauthor{\bsnm{Bairoch}, \binits{A.}},
\bauthor{\bsnm{Apweiler}, \binits{R.}}:
\batitle{The swiss-prot protein sequence database and its supplement trembl in
  2000}.
\bjtitle{Nucleic acids research}
\bvolume{28}(\bissue{1}),
\bfpage{45}--\blpage{48}
(\byear{2000})
\end{barticle}
\endbibitem

\bibitem[\protect\citeauthoryear{Mistry et~al.}{2021}]{mistry2021pfam}
\begin{barticle}
\bauthor{\bsnm{Mistry}, \binits{J.}},
\bauthor{\bsnm{Chuguransky}, \binits{S.}},
\bauthor{\bsnm{Williams}, \binits{L.}},
\bauthor{\bsnm{Qureshi}, \binits{M.}},
\bauthor{\bsnm{Salazar}, \binits{G.A.}},
\bauthor{\bsnm{Sonnhammer}, \binits{E.L.}},
\bauthor{\bsnm{Tosatto}, \binits{S.C.}},
\bauthor{\bsnm{Paladin}, \binits{L.}},
\bauthor{\bsnm{Raj}, \binits{S.}},
\bauthor{\bsnm{Richardson}, \binits{L.J.}}, \betal:
\batitle{Pfam: The protein families database in 2021}.
\bjtitle{Nucleic acids research}
\bvolume{49}(\bissue{D1}),
\bfpage{412}--\blpage{419}
(\byear{2021})
\end{barticle}
\endbibitem

\bibitem[\protect\citeauthoryear{Paysan-Lafosse et~al.}{2025}]{paysan2025pfam}
\begin{barticle}
\bauthor{\bsnm{Paysan-Lafosse}, \binits{T.}},
\bauthor{\bsnm{Andreeva}, \binits{A.}},
\bauthor{\bsnm{Blum}, \binits{M.}},
\bauthor{\bsnm{Chuguransky}, \binits{S.R.}},
\bauthor{\bsnm{Grego}, \binits{T.}},
\bauthor{\bsnm{Pinto}, \binits{B.L.}},
\bauthor{\bsnm{Salazar}, \binits{G.A.}},
\bauthor{\bsnm{Bileschi}, \binits{M.L.}},
\bauthor{\bsnm{Llinares-L{\'o}pez}, \binits{F.}},
\bauthor{\bsnm{Meng-Papaxanthos}, \binits{L.}}, \betal:
\batitle{The pfam protein families database: embracing ai/ml}.
\bjtitle{Nucleic acids research}
\bvolume{53}(\bissue{D1}),
\bfpage{523}--\blpage{534}
(\byear{2025})
\end{barticle}
\endbibitem

\bibitem[\protect\citeauthoryear{Xu et~al.}{2018}]{xu2018powerful}
\begin{botherref}
\oauthor{\bsnm{Xu}, \binits{K.}},
\oauthor{\bsnm{Hu}, \binits{W.}},
\oauthor{\bsnm{Leskovec}, \binits{J.}},
\oauthor{\bsnm{Jegelka}, \binits{S.}}:
How powerful are graph neural networks?
arXiv preprint arXiv:1810.00826
(2018)
\end{botherref}
\endbibitem

\bibitem[\protect\citeauthoryear{Zhang et~al.}{2022}]{zhang2022use}
\begin{bchapter}
\bauthor{\bsnm{Zhang}, \binits{S.}},
\bauthor{\bsnm{Xu}, \binits{R.}},
\bauthor{\bsnm{Xiong}, \binits{C.}},
\bauthor{\bsnm{Ramaiah}, \binits{C.}}:
\bctitle{Use all the labels: A hierarchical multi-label contrastive learning
  framework}.
In: \bbtitle{Proceedings of the IEEE/CVF Conference on Computer Vision and
  Pattern Recognition},
pp. \bfpage{16660}--\blpage{16669}
(\byear{2022})
\end{bchapter}
\endbibitem

\bibitem[\protect\citeauthoryear{Brandes et~al.}{2022}]{brandes2022proteinbert}
\begin{barticle}
\bauthor{\bsnm{Brandes}, \binits{N.}},
\bauthor{\bsnm{Ofer}, \binits{D.}},
\bauthor{\bsnm{Peleg}, \binits{Y.}},
\bauthor{\bsnm{Rappoport}, \binits{N.}},
\bauthor{\bsnm{Linial}, \binits{M.}}:
\batitle{Proteinbert: a universal deep-learning model of protein sequence and
  function}.
\bjtitle{Bioinformatics}
\bvolume{38}(\bissue{8}),
\bfpage{2102}--\blpage{2110}
(\byear{2022})
\end{barticle}
\endbibitem

\bibitem[\protect\citeauthoryear{Jin et~al.}{2019}]{jin2019probing}
\begin{botherref}
\oauthor{\bsnm{Jin}, \binits{Q.}},
\oauthor{\bsnm{Dhingra}, \binits{B.}},
\oauthor{\bsnm{Cohen}, \binits{W.W.}},
\oauthor{\bsnm{Lu}, \binits{X.}}:
Probing biomedical embeddings from language models.
arXiv preprint arXiv:1904.02181
(2019)
\end{botherref}
\endbibitem

\bibitem[\protect\citeauthoryear{van~den Oord
  et~al.}{2018}]{journals/corr/abs-1807-03748}
\begin{botherref}
\oauthor{\bsnm{Oord}, \binits{A.}},
\oauthor{\bsnm{Li}, \binits{Y.}},
\oauthor{\bsnm{Vinyals}, \binits{O.}}:
Representation learning with contrastive predictive coding.
CoRR
\textbf{abs/1807.03748}
(2018)
\end{botherref}
\endbibitem

\bibitem[\protect\citeauthoryear{Rives et~al.}{2019}]{rives2019biological}
\begin{barticle}
\bauthor{\bsnm{Rives}, \binits{A.}},
\bauthor{\bsnm{Meier}, \binits{J.}},
\bauthor{\bsnm{Sercu}, \binits{T.}},
\bauthor{\bsnm{Goyal}, \binits{S.}},
\bauthor{\bsnm{Lin}, \binits{Z.}},
\bauthor{\bsnm{Liu}, \binits{J.}},
\bauthor{\bsnm{Guo}, \binits{D.}},
\bauthor{\bsnm{Ott}, \binits{M.}},
\bauthor{\bsnm{Zitnick}, \binits{C.L.}},
\bauthor{\bsnm{Ma}, \binits{J.}},
\bauthor{\bsnm{Fergus}, \binits{R.}}:
\batitle{Biological structure and function emerge from scaling unsupervised
  learning to 250 million protein sequences}.
\bjtitle{PNAS}
(\byear{2019})
\doiurl{10.1101/622803}
\end{barticle}
\endbibitem

\bibitem[\protect\citeauthoryear{Szymborski and Emad}{2024}]{intrepppid}
\begin{barticle}
\bauthor{\bsnm{Szymborski}, \binits{J.}},
\bauthor{\bsnm{Emad}, \binits{A.}}:
\batitle{Intrepppid—an orthologue-informed quintuplet network for
  cross-species prediction of protein–protein interaction}.
\bjtitle{Briefings in Bioinformatics}
\bvolume{25}(\bissue{5}),
\bfpage{405}
(\byear{2024})
\doiurl{10.1093/bib/bbae405}
\end{barticle}
\endbibitem

\end{thebibliography}
\end{document}